\documentclass[runningheads]{llncs}

\usepackage{eccv}

\usepackage{eccvabbrv}

\usepackage[accsupp]{axessibility}  

\usepackage{cite}
\usepackage{comment}
\usepackage{amsmath,amssymb,amsfonts}
\usepackage{algorithmic}
\usepackage{graphicx}
\usepackage{textcomp}
\usepackage{booktabs}
\usepackage{url}
\usepackage[table]{xcolor}
\usepackage{float}
\usepackage{ulem}
\usepackage{multirow}
\usepackage{orcidlink}

\usepackage{hyperref}

\newcommand{\std}[1]{\textcolor{gray}{\scriptsize$\pm$ #1}}

\definecolor{lightblue}{RGB}{230,242,255}

\begin{document}


\title{SLAP: Selective Local Vision-Language Alignment for Fish
Re-Identification via Partial Optimal Transport}

\titlerunning{SLAP for Fish Re-Identification}

\author{
Cigdem Beyan\inst{1}\orcidlink{0000-0002-9583-0087}\thanks{Corresponding author.}
\and
Tonje Knutsen S{\o}rdalen\inst{2,3}\orcidlink{0000-0001-5836-9327}
\and
Kim Tallaksen Halvorsen\inst{2}\orcidlink{0000-0001-6857-2492}
}

\authorrunning{C.~Beyan et al.}

\institute{
Department of Computer Science, University of Verona,
37134 Verona, Italy
\email{cigdem.beyan@univr.it}
\and
Institute of Marine Research, Nye Fl{\o}devigveien 20,
4817 His, Norway
\email{\{tonje.sordalen,kim.halvorsen\}@hi.no}
\and
University of Agder, Centre for Coastal Research,
4604 Grimstad, Norway
}

\maketitle

\begin{abstract}
Individual fish re-identification (ReID) is a fine-grained recognition problem in which identity-discriminative cues are often localized to specific body regions rather than distributed uniformly across the animal. Nevertheless, recent CLIP-based ReID methods rely predominantly on global image-text alignment, allowing background and weakly discriminative regions to contribute to cross-modal supervision. We propose a selective local vision-language alignment framework that establishes localized correspondences between visual patch embeddings and multiple identity-aware prompt embeddings through Partial Optimal Transport (POT). Rather than enforcing exhaustive correspondence, POT enables selective matching between visual patches and prompt embeddings, allowing the model to emphasize the strongest cross-modal correspondences while avoiding forced alignment of weakly matching regions, thereby yielding more discriminative visual representations for retrieval. The framework is trained end-to-end, while only the adapted visual encoder is retained during inference. Experiments on the longitudinal Symphodus melops dataset demonstrate consistent improvements over recent CLIP-based ReID methods under both closed-set and open-set evaluation protocols. Additional evaluations on other datasets further demonstrate the generalization capability of the proposed method across diverse marine ReID benchmarks.


\end{abstract}

\begin{figure*}[!h]
    \centering
    \includegraphics[width=\textwidth]{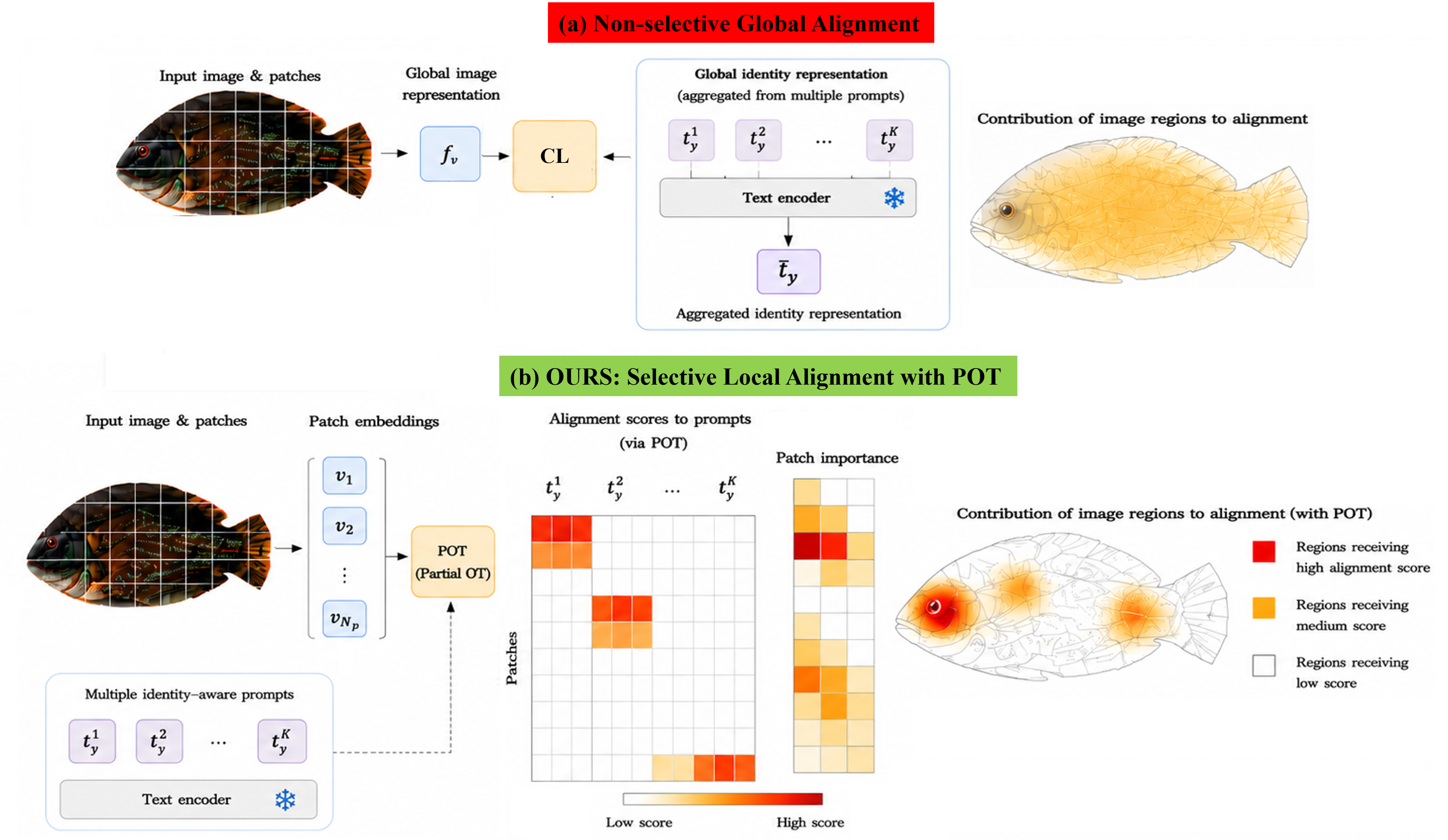}
    \caption{Comparison between conventional global alignment and our SLAP. (a) The image encoder produces a single global visual representation aligned with an aggregated identity-aware textual representation. Since alignment is performed globally, all image regions contribute similarly to the matching process, including weakly discriminative or background regions. (b) SLAP instead aligns local visual patches with multiple identity-aware prompt embeddings through POT. Resulting transport plan emphasizes regions receiving strong cross-modal correspondence while suppressing weakly aligned regions.}
    \label{fig:potTeaser}
\end{figure*}

\section{Introduction}
\label{sec:intro}

Animal re-identification (ReID) refers to recognizing previously observed individuals across repeated encounters and is an important component of wildlife monitoring, behavioral ecology, and longitudinal population studies~\cite{mcclintock2014mark}. 
In computer vision, animal ReID specifically denotes identifying individuals from images or video sequences without requiring invasive physical tagging during inference~\cite{huang2025uniformity,compte2025housed,liu2024fishtrack}. 
Automated visual ReID therefore requires models to learn discriminative identity representations capable of matching the same individual despite substantial appearance variability across observations acquired under natural and unconstrained conditions~\cite{nepovinnykh2020siamese,ravoor2020survey,li2020atrw,Beyan2026}.

Animal ReID remains particularly challenging due to long-term temporal variation and subtle inter-individual differences~\cite{ravoor2020survey,Beyan2026}.
Animals exhibit deformable body structures, large pose variability, occlusions, and fine-grained identity-specific patterns that are often difficult to distinguish visually. 
In addition, appearance may evolve substantially over time because of growth, aging, seasonal coloration changes, or life-stage transitions, increasing intra-individual variation and complicating identity matching across repeated observations spanning months or years~\cite{cermak2024wildfusion,ravoor2020survey,jiao2023toward,Beyan2026}.

In this work, we investigate fish ReID using the Melops dataset~\cite{sordalen2025melopsreid,Sordalen2026wild}, a longitudinal benchmark collected from wild corkwing wrasse (\textit{Symphodus melops}). 
The dataset captures realistic long-term appearance variation under in-the-wild conditions, where individuals may undergo substantial changes in body size, pigmentation, and overall appearance across repeated observations spanning months or years. In addition, strong left-right asymmetry in fish pigmentation and body patterns further increases intra-individual variability, while the identity distribution remains highly imbalanced because many individuals are observed only once. These properties make Melops a particularly challenging benchmark for studying long-term fish ReID and temporally robust identity representation learning.

Recent animal ReID approaches have increasingly adopted vision-language models (VLMs), particularly CLIP~\cite{radford2021learning}, to leverage transferable representations learned from image-text supervision. 
These methods adapt pretrained vision-language encoders through prompt learning, cross-modal alignment, and parameter-efficient fine-tuning strategies~\cite{li2023clip,wu2024individual,turESWA2026,jiao2023toward,li2025metawild}. 
For example, CLIP-ReID~\cite{li2023clip} introduces learnable identity-aware prompts together with visual adaptation modules to transfer pretrained vision-language knowledge into the ReID setting. 
Similarly, IndivAID~\cite{wu2024individual} employs individual-aware textual descriptions to guide visual representation learning through cross-modal supervision. More recently, metadata-aware vision-language adaptation has also been explored for longitudinal animal ReID~\cite{turESWA2026}, where biological metadata, learnable prompts, and parameter-efficient visual adaptation are jointly incorporated to improve robustness against temporal appearance variation.

Despite their strong performance, existing CLIP-based animal ReID approaches still rely primarily on \textit{global image-level alignment} between visual and textual representations. CLIP-ReID~\cite{li2023clip} and IndivAID~\cite{wu2024individual} both employ multi-stage optimization strategies in which textual representations are first constructed or optimized before subsequently guiding visual adaptation through cross-modal supervision. Similarly, MetaPrompt-ReID~\cite{turESWA2026} incorporates learnable prompts, metadata conditioning, and parameter-efficient visual adaptation within a global alignment framework. However, under global alignment, textual supervision mainly acts as a coarse semantic constraint while the visual branch continues to dominate the representation learning process. As a result, local identity-discriminative regions may contribute weakly to the cross-modal interaction, while background regions and weakly discriminative body parts can still influence the learned representation. This limitation becomes particularly important in longitudinal fish ReID, where identity-relevant cues are often localized and may evolve unevenly across body regions over time. Biological studies further suggest that stable identity-related cues in fish frequently concentrate around localized morphological structures such as the operculum, eye, and snout regions~\cite{ellis_visual_2026}. This observation is additionally supported by the findings in MetaPrompt-ReID~\cite{turESWA2026}, where head-only and full-body experiments exhibit different retrieval performance, indicating that identity-discriminative information is not uniformly distributed across the fish body. These observations motivate the need for a more selective cross-modal alignment mechanism capable of establishing \textit{localized prompt-to-patch correspondences} while suppressing weak or noisy matches during representation learning.

Consequently, we propose a vision-language adaptation framework for longitudinal animal ReID based on \textit{selective local vision-language alignment}, called \textbf{SLAP}. Instead of relying solely on global image-text correspondence, SLAP establishes localized interactions between visual patch embeddings and multiple identity-aware prompt embeddings through Partial Optimal Transport (POT)~\cite{villani2008optimal}. This formulation enables the model to selectively emphasize informative local regions while potentially suppressing weakly aligned or noisy correspondences during representation learning (See Fig.~\ref{fig:potTeaser}). In addition, SLAP combines parameter-efficient visual adaptation and learnable prompt tuning to preserve transferable pretrained representations while adapting the model to the fine-grained and temporally varying nature of fish identities. Experimental results on the Melops dataset demonstrate that SLAP consistently improves ReID performance across various evaluation protocols compared to state-of-the-art (SOTA) methods. Additional evaluations on multiple marine animal ReID datasets further demonstrate the robustness and generalization capability of our method across diverse species and acquisition conditions.

The main contributions are threefold: (1) We introduce a selective local vision-language alignment formulation for animal ReID that establishes prompt-to-patch correspondences through POT, replacing conventional global image-text alignment. (2) We develop a lightweight vision-language adaptation framework that integrates the proposed POT formulation with learnable identity-aware prompts while retaining a purely visual inference pipeline. (3) Our SLAP improves performance on the challenging Melops dataset and further demonstrates strong generalization on additional marine animal ReID benchmarks.

\section{Related Work}
\label{sec:realted}

\noindent\textbf{Vision-Language Adaptation for Animal ReID.}
Recent animal ReID approaches have increasingly adopted VLMs, to exploit transferable representations learned from large-scale image-text supervision and provide cross-modal guidance even when only identity labels are available. CLIP-ReID~\cite{li2023clip} first adapts CLIP to ReID through learnable identity-aware prompts, but relies on staged optimization in which prompt learning and visual adaptation are performed sequentially. IndivAID~\cite{wu2024individual} extends this idea by generating image- and individual-specific textual representations to guide visual feature learning, introducing additional modules and a multi-stage training pipeline. Jiao et al.~\cite{jiao2023toward} further incorporate large language models (LLMs) to construct semantic prompts for cross-species and open-world animal ReID, increasing model complexity and dependence on external prompt generation. Beyond language-only supervision, MetaWild~\cite{li2025metawild} incorporates environmental metadata through adapter-style fusion modules, while MetaPrompt-ReID~\cite{turESWA2026} introduces metadata-aware vision-language adaptation through learnable prompts, metadata conditioning, and parameter-efficient visual adaptation. Despite these advances, existing methods continue to rely primarily on global image-level alignment between visual and textual representations, while auxiliary information is typically incorporated through discrete textual conversion or additional fusion modules, increasing architectural complexity and limiting localized cross-modal interaction.

\noindent\textbf{Parameter-Efficient Fine-Tuning \& Prompt Learning.}
PEFT adapts large pretrained models by updating only a small subset of parameters, reducing computational cost and overfitting compared with full fine-tuning~\cite{houlsby2019parameter,zaken2022bitfit}. Representative approaches include adapter modules, bias-only tuning, prompt learning, and low-rank adaptation. Among these, LoRA~\cite{lora2022} has become one of the most widely adopted methods by introducing trainable low-rank updates without increasing inference complexity. Other lightweight adaptation strategies include residual adapters~\cite{houlsby2019parameter}, visual prompt tuning~\cite{jia2022visual}, and Tip-Adapter~\cite{zhang2022tip}. Prompt learning provides a complementary adaptation mechanism by optimizing continuous prompts while keeping the backbone frozen. CoOp~\cite{zhou2022coop} first demonstrated the effectiveness of this strategy for vision--language adaptation, with subsequent works extending it through class-specific prompts, deeper prompt insertion, and hybrid prompt-adapter formulations~\cite{zhou2022coop,jia2022visual}. Since the effectiveness of PEFT strategies is task-dependent, we adopt LoRA following the comparative analysis in MetaPrompt-ReID~\cite{turESWA2026}, where it consistently outperformed alternative adaptation methods on the Melops benchmark while maintaining parameter efficiency. \\

\noindent\textbf{Optimal Transport.} OT~\cite{villani2008optimal,peyre2019computational} is for modeling correspondences between probability distributions and has found applications in domain adaptation~\cite{li2024unsupervised}, noisy-label learning~\cite{feng2023ot}, causal inference~\cite{tu2022optimal}, anomaly detection~\cite{shiri2025madpot}, and federated learning~\cite{li2024global}. Partial OT extends OT by relaxing full mass preservation, enabling selective matching when only a subset of elements should participate in the alignment. Within vision-language learning, PLOT~\cite{chen2022plot} introduces OT-based local prompt learning for fine-grained image-text alignment. However, the application of OT and POT to localized vision-language alignment for animal ReID under long-term appearance variation remains largely unexplored.

\section{Proposed Method}
\label{sec:method}

\begin{figure*}[!tb]
    \centering
    \includegraphics[width=\textwidth]{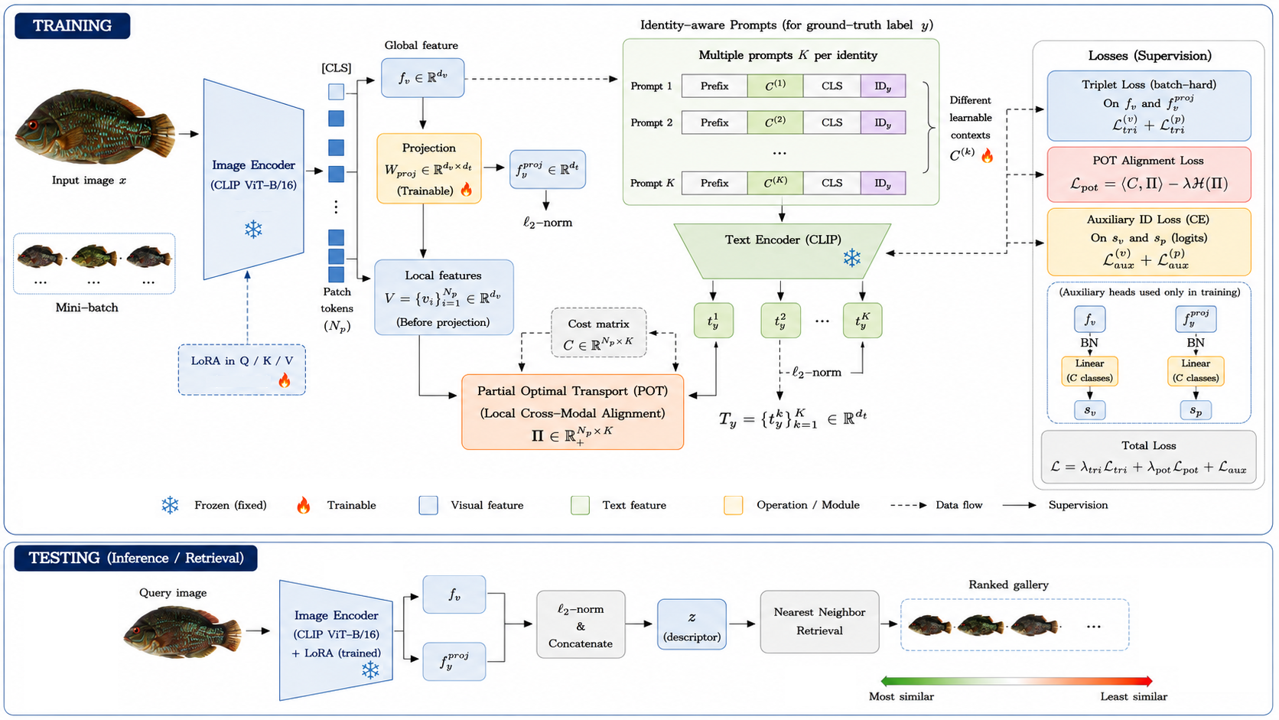}
    \caption{Overview of SLAP. A frozen CLIP image encoder is adapted using LoRA modules inserted into the query, key, and value projection layers of the transformer attention blocks. The image encoder produces both global visual representations $(\mathbf{f}_v,\mathbf{f}_v^{\mathrm{proj}})$ and local patch embeddings $\mathbf{V}$. Identity-aware prompts composed of fixed tokens and learnable context tokens are processed by a frozen CLIP text encoder to produce multiple prompt embeddings $\mathbf{T}_y$. POT establishes selective local cross-modal alignment between visual patch embeddings and prompt embeddings through a transport plan $\mathbf{\Pi}$. SLAP is jointly optimized using batch-hard triplet losses, POT-based alignment loss, and auxiliary identity classification losses. During inference, only the LoRA-adapted image encoder is retained. The normalized global and projected visual representations are concatenated and re-normalized to form the final descriptor used for nearest-neighbor retrieval.}
    \label{fig:proposed}
\end{figure*}

Given a training set of image-identity pairs
$\{(\mathbf{x}_i,y_i)\}_{i=1}^{N}$,
where $\mathbf{x}_i\in\mathbb{R}^{H\times W\times3}$ denotes an input image and $y_i\in\{1,\ldots,C\}$ the corresponding identity label, the objective of animal Re-ID is to learn an embedding function
$f:\mathcal{X}\rightarrow\mathbb{R}^{d}$
that maps images of the same individual close together while separating different identities in the learned feature space. To this end, we propose SLAP, a vision-language adaptation framework that combines learnable identity-aware prompts with selective local cross-modal alignment through POT~\cite{phatak2023computing}. Unlike conventional global alignment, SLAP establishes localized correspondences between local visual embeddings and multiple identity-aware prompt embeddings, allowing only a subset of visual patches to participate in the alignment process. This formulation reduces the influence of background clutter, pose changes, and other non-informative visual patterns while capturing complementary identity-related appearance cues through multiple prompt embeddings. The text branch is used exclusively during training to provide structured cross-modal supervision for learning discriminative visual representations. At inference time, only the visual representations are retained for nearest-neighbor retrieval, resulting in a purely visual deployment pipeline. An overview of the proposed framework is shown in Fig.~\ref{fig:proposed}. \\

\noindent\textbf{Visual Feature and Prompt Encoding.}
SLAP employs frozen transformer-based image and text encoders to preserve the transferable vision-language knowledge learned during large-scale pretraining. The visual branch is adapted through Low-Rank Adaptation (LoRA)~\cite{lora2022}, while the text branch is adapted using learnable prompt tokens. 

Given an input image $\mathbf{x}$, the image encoder produces a sequence of visual token embeddings from which both a global representation $\mathbf{f}_v$ and local patch embeddings $\mathbf{V}=\{\mathbf{v}_1,\ldots,\mathbf{v}_{N_p}\}$ are obtained, where $N_p$ denotes the number of visual patches. The global representation is projected into the shared vision-language embedding space as $\mathbf{f}_v^{\mathrm{proj}}=\mathbf{f}_v\mathbf{W}_{\mathrm{proj}}$, where $\mathbf{W}_{\mathrm{proj}}\in\mathbb{R}^{d_v\times d_t}$ maps the visual features to the text embedding dimension $d_t$. The projected global embeddings are $\ell_2$-normalized before metric learning. 

On the text side, each identity $y$ is associated with a set of prompts $\mathcal{P}_y=\{p_y^1,\ldots,p_y^K\}$, where $K$ denotes the number of prompt instances. Each prompt consists of fixed template tokens and learnable context tokens, producing identity-aware prompt embeddings $\mathbf{t}_y^k=\mathrm{TextEncoder}(p_y^k)$. The resulting prompt embeddings are grouped as $\mathbf{T}_y=\{\mathbf{t}_y^1,\ldots,\mathbf{t}_y^K\}$. Since both the local visual patch embeddings $\mathbf{V}$ and the prompt embeddings $\mathbf{T}_y$ are extracted from the pretrained CLIP encoders, they naturally reside in the shared vision-language embedding space. 

SLAP performs selective local cross-modal alignment between $\mathbf{V}$ and $\mathbf{T}_y$ through POT, producing a transport plan that establishes soft prompt-to-patch correspondences for cross-modal supervision during training. \\

\noindent \textbf{Low-Rank Adaptation.}
Rather than updating the full set of encoder parameters, LoRA~\cite{lora2022} introduces trainable low-rank residual updates into the attention projection layers of the transformer architecture. 
In particular, the adaptation is applied to the query, key, and value projections within each multi-head self-attention block.
Let $\mathbf{W}_0 \in \mathbb{R}^{d_{\mathrm{out}} \times d_{\mathrm{in}}}$ denote a pretrained linear projection matrix. 
LoRA parameterizes the adaptation as a low-rank decomposition added to the original projection, modifying the forward projection as $\mathbf{h} = \mathbf{W}_0 \mathbf{u} + \frac{\alpha}{r}\mathbf{B}\mathbf{A}\mathbf{u}$, where $\mathbf{A} \in \mathbb{R}^{r \times d_{\mathrm{in}}}$ and $\mathbf{B} \in \mathbb{R}^{d_{\mathrm{out}} \times r}$ are trainable low-rank matrices with rank $r \ll \min(d_{\mathrm{out}}, d_{\mathrm{in}})$. 
The scalar $\alpha$ scales the contribution of the low-rank update during training. 
Equivalently, the adapted projection matrix can be expressed as $\mathbf{W} = \mathbf{W}_0 + \frac{\alpha}{r}\mathbf{B}\mathbf{A}$.
During optimization, only the low-rank matrices are updated, while the pretrained encoder parameters remain fixed. 
This enables lightweight adaptation of the visual representations while maintaining the general semantic knowledge encoded in the pretrained backbone. \\

\noindent\textbf{Learnable Prompt-Based Text Representation.}
To adapt the frozen text encoder to the animal ReID task, SLAP employs learnable prompt tuning inspired by CoOp~\cite{zhou2022coop}. This adaptation is achieved through learnable context tokens inserted into a fixed textual template. For each identity $y$, the prompt set $\mathcal{P}_y=\{p_y^1,\ldots,p_y^K\}$ introduced above is parameterized by an independent set of learnable context tokens together with a fixed identity-specific token embedding. Unlike proxy representations commonly used in metric learning, these identity tokens exist exclusively in the text embedding space and are used only during training. The identity token is shared across all samples belonging to the same identity, remains fixed throughout training, and provides identity-aware conditioning of the prompt representations. 

To improve robustness to prompt structure, SLAP adopts three prompt layouts following~\cite{turESWA2026,zhou2022coop}: end, middle, and front, obtained by varying the placement of the learnable context tokens relative to the fixed template tokens. Specifically, the layouts are defined as $[\mathbf{e}_{\mathrm{prefix}},\,\mathbf{e}_{\mathrm{cls}},\,\mathbf{C}^{(k)},\,\mathbf{e}_{\mathrm{id}}^{\,y}]$, $[\mathbf{e}_{\mathrm{prefix}},\,\mathbf{C}^{(k)}_1,\,\mathbf{e}_{\mathrm{cls}},\,\mathbf{C}^{(k)}_2,\,\mathbf{e}_{\mathrm{id}}^{\,y}]$, and $[\mathbf{e}_{\mathrm{prefix}},\,\mathbf{C}^{(k)},\,\mathbf{e}_{\mathrm{cls}},\,\mathbf{e}_{\mathrm{id}}^{\,y}]$, where $\mathbf{C}^{(k)}\in\mathbb{R}^{M\times d_t}$ denotes the $M$ learnable context tokens and $\mathbf{C}^{(k)}_1,\mathbf{C}^{(k)}_2$ correspond to their first and second halves. The three layouts share the same learnable context tokens and identity token, producing alternative token arrangements of the same prompt rather than distinct prompts. The resulting embeddings are averaged after $\ell_2$ normalization, yielding one embedding per prompt instance. In training, only the learnable context tokens are optimized jointly with the LoRA parameters, while the pretrained text encoder, identity tokens, and fixed template tokens remain frozen. The resulting prompt embeddings provide identity-aware cross-modal supervision and are discarded in inference. \\

\noindent\textbf{Partial Optimal Transport.}
Unlike conventional global alignment strategies that establish a single correspondence between an image and a text representation, SLAP employs POT to perform selective local alignment between visual patch embeddings and multiple identity-aware prompt embeddings. 

Given local visual embeddings $\mathbf{V}=\{\mathbf{v}_1,\ldots,\mathbf{v}_{N_p}\}$ and prompt embeddings $\mathbf{T}_y=\{\mathbf{t}_y^1,\ldots,\mathbf{t}_y^K\}$, we construct a transport cost matrix $\mathbf{C}\in\mathbb{R}^{N_p\times K}$ in the shared vision-language embedding space, where each element is defined as the cosine distance
$
\mathbf{C}_{ik}
=
1-
\frac{\mathbf{v}_i^\top\mathbf{t}_y^k}
{\|\mathbf{v}_i\|_2\|\mathbf{t}_y^k\|_2}.
$
Lower transport costs therefore correspond to stronger visual-text similarity. Let $\boldsymbol{\mu}\in\mathbb{R}^{N_p}$ and $\boldsymbol{\nu}\in\mathbb{R}^{K}$ denote uniform source and target distributions over visual patches and prompt embeddings, respectively. The transport plan is represented by a non-negative matrix $\mathbf{\Pi}\in\mathbb{R}_{+}^{N_p\times K}$, where $\mathbf{\Pi}_{ik}$ denotes the transported mass between $\mathbf{v}_i$ and $\mathbf{t}_y^k$. Following~\cite{villani2008optimal,phatak2023computing}, the transport problem is formulated as
$
\min_{\mathbf{\Pi}\ge0}
\langle\mathbf{C},\mathbf{\Pi}\rangle,
$
subject to the relaxed marginal constraints
$
\mathbf{\Pi}\mathbf{1}_{K}\le\boldsymbol{\mu},
$
$
\mathbf{\Pi}^{\top}\mathbf{1}_{N_p}\le\boldsymbol{\nu},
$
and the partial mass constraint
$
\mathbf{1}_{N_p}^{\top}\mathbf{\Pi}\mathbf{1}_{K}=m,
$
where
$
m=\rho\min(\|\boldsymbol{\mu}\|_1,\|\boldsymbol{\nu}\|_1),
$
$\rho\in(0,1]$ denotes the transported mass ratio, $\langle\cdot,\cdot\rangle$ is the Frobenius inner product, and $\mathbf{1}_d$ denotes a $d$-dimensional vector of ones. Compared with standard OT, the relaxed constraints allow only a subset of visual patches to participate in the alignment process. To improve optimization stability, we employ entropic regularization, yielding
$
\min_{\mathbf{\Pi}\ge0}
\langle\mathbf{C},\mathbf{\Pi}\rangle
-\lambda\mathcal{H}(\mathbf{\Pi}),
$
where
$
\mathcal{H}(\mathbf{\Pi})
=
-\sum_{i,k}\mathbf{\Pi}_{ik}\log\mathbf{\Pi}_{ik}
$
is the entropy regularizer and $\lambda$ controls its strength. In training, the transport objective is computed between the local visual embeddings of each sample and the prompt embeddings associated with its ground-truth identity. The resulting transport plan establishes soft prompt-to-patch correspondences that provide selective local cross-modal supervision. \\

\noindent\textbf{Auxiliary Identity Supervision.}
Although the prompt embeddings provide identity-aware cross-modal supervision, they do not explicitly enforce identity separation within the visual embedding space. To further improve discriminative feature learning, auxiliary identity supervision is applied to both the global visual representation $\mathbf{f}_v\in\mathbb{R}^{d_v}$ and its projected counterpart $\mathbf{f}_v^{\mathrm{proj}}\in\mathbb{R}^{d_t}$. Specifically,
$\mathbf{s}_v=\mathbf{W}_{\mathrm{aux}}^{(v)}\mathrm{BN}(\mathbf{f}_v)$ and
$\mathbf{s}_p=\mathbf{W}_{\mathrm{aux}}^{(p)}\mathrm{BN}(\mathbf{f}_v^{\mathrm{proj}})$,
where $\mathbf{W}_{\mathrm{aux}}^{(v)}\in\mathbb{R}^{C\times d_v}$ and
$\mathbf{W}_{\mathrm{aux}}^{(p)}\in\mathbb{R}^{C\times d_t}$ denote trainable classification matrices. The resulting logits are supervised using the cross-entropy loss with identity labels. Supervising both feature spaces promotes identity discrimination before and after projection into the shared vision-language embedding space, stabilizing optimization and improving feature separability. The auxiliary classification heads are used only during training and discarded during inference. \\

\noindent\textbf{Loss Function.} SLAP is optimized using a composite objective that combines metric learning, POT-based cross-modal alignment, and auxiliary identity supervision. The POT objective serves as the cross-modal alignment loss, $\mathcal{L}_{\mathrm{pot}}=\langle\mathbf{C},\mathbf{\Pi}\rangle-\lambda\mathcal{H}(\mathbf{\Pi})$, where $\mathbf{C}$ denotes the transport cost matrix, $\mathbf{\Pi}$ the optimal transport plan, $\mathcal{H}(\mathbf{\Pi})=-\sum_{i,k}\mathbf{\Pi}_{ik}\log\mathbf{\Pi}_{ik}$ the entropy regularization term, and $\lambda$ its weighting factor. Batch-hard triplet losses are independently applied to the global image representation $\mathbf{f}_v$ and its projected counterpart $\mathbf{f}_v^{\mathrm{proj}}$, yielding $\mathcal{L}_{\mathrm{tri}}=\mathcal{L}_{\mathrm{tri}}^{(v)}(\mathbf{f}_v,y)+\mathcal{L}_{\mathrm{tri}}^{(p)}(\mathbf{f}_v^{\mathrm{proj}},y)$. Triplet supervision is computed independently within each embedding space using $\ell_2$-normalized embeddings, while Euclidean distance is employed for both optimization and retrieval. Auxiliary identity supervision is applied to the classification logits $\mathbf{s}_v$ and $\mathbf{s}_p$ through the cross-entropy loss, $\mathcal{L}_{\mathrm{aux}}=\mathcal{L}_{\mathrm{CE}}(\mathbf{s}_v,y)+\mathcal{L}_{\mathrm{CE}}(\mathbf{s}_p,y)$. The overall objective is defined as $\mathcal{L}=\lambda_{\mathrm{tri}}\mathcal{L}_{\mathrm{tri}}+\lambda_{\mathrm{pot}}\mathcal{L}_{\mathrm{pot}}+\mathcal{L}_{\mathrm{aux}}$, where $\lambda_{\mathrm{tri}}$ and $\lambda_{\mathrm{pot}}$ balance the contributions of metric learning and POT-based cross-modal alignment, respectively. During optimization, only the LoRA parameters, learnable context tokens, auxiliary classification heads, and normalization layers are updated, while the pretrained image and text encoders remain frozen. \\

\noindent\textbf{Inference.}
Only the LoRA-adapted image encoder is retained, while all text-related components, prompt representations, POT-based alignment, and auxiliary supervision heads are discarded. Given a query image and a gallery set, ReID is performed through nearest-neighbor retrieval using Euclidean distance between visual descriptors. Specifically, the global image representation $\mathbf{f}_v$ and its projected counterpart $\mathbf{f}_v^{\mathrm{proj}}$ are individually $\ell_2$-normalized, concatenated as $[\mathrm{norm}(\mathbf{f}_v);\mathrm{norm}(\mathbf{f}_v^{\mathrm{proj}})]$, and normalized once more to obtain the final descriptor $\mathbf{z}=\mathrm{norm}([\mathrm{norm}(\mathbf{f}_v);\mathrm{norm}(\mathbf{f}_v^{\mathrm{proj}})])$. The fusion combines complementary information from the original visual and shared vision-language embedding spaces without introducing additional trainable parameters. Following~\cite{turESWA2026}, fish Re-ID is formulated as an image retrieval problem rather than a closed-set identity classification task, enabling generalization to previously unseen identities without requiring a fixed identity label space during inference.

\section{Experimental Analysis}
\label{sec:experiments}

\noindent\textbf{Melops Dataset.}
This is~\cite{sordalen2025melopsreid,Sordalen2026wild}, a large-scale longitudinal dataset for individual ReID of wild corkwing wrasse (\textit{Symphodus melops}). It contains 24578 images of 9861 PIT-tagged individuals collected between 2018 and 2024 in Western Norway under a long-term capture-mark-recapture program. Each fish was photographed from both lateral sides under standardized imaging conditions using a uniform background and color reference card. Among the recorded individuals, 1883 were recaptured at least once, resulting in 2916 resighting events (8524 images) spanning intervals from within-season recaptures to multiple years. These repeated observations introduce substantial variation in body size, pigmentation, and appearance, while the majority of individuals are observed only once, resulting in a highly imbalanced identity distribution. Furthermore, \textit{Symphodus melops} exhibits pronounced left-right asymmetry in pigmentation and body patterns, increasing intra-individual appearance variation and preventing trivial matching between mirrored views. \\

\noindent\textbf{Evaluation Protocols.}
The evaluation is performed under two protocols differing in the identity overlap between the training and test sets. In the \textbf{closed-set} (CS) protocol, all test identities are observed during training, while training and testing images remain disjoint. Approximately 70\% of the images of each identity are assigned to training, and the remainder is divided into query and gallery sets following standard ReID practice~\cite{li2020atrw,li2023clip,wu2021deep,jiao2023toward,cermak2024wildfusion,adam2024seaturtleid}. Identities with fewer than three samples are assigned entirely to the training split, resulting in 9814 training identities (20656 images), 1843 query images, and 1953 gallery images. Following~\cite{adam2024seaturtleid}, the \textbf{open-set} (OS) protocol additionally evaluates generalization to unseen identities. Approximately 20\% of the test identities are excluded from training, while the remaining identities overlap with the training split. Query and gallery sets contain both seen and unseen identities, resulting in 9470 training identities (19761 images), 2254 query images, and 2533 gallery images. \\

\noindent \textbf{Evaluation Metrics.}
ReID performance is evaluated using standard retrieval metrics based on Euclidean distances between image embeddings. 
For each query, gallery samples are ranked according to ascending distance in the learned embedding space. 
\textbf{Cumulative Matching Characteristic (CMC)} measures whether at least one correct match appears within the top-\(k\) retrieved gallery samples. We report \textbf{Rank-1} and \textbf{Rank-5 accuracy} in line with earlier methods. \textbf{Mean Average Precision (mAP)} accounts for multiple correct matches per query and is computed as the mean of the average precision values across all queries \cite{ravoor2020survey,Beyan2026}. \\

\noindent\textbf{Implementation Details.} We employ a CLIP ViT-B/16 backbone~\cite{radford2021learning} initialized from publicly available pretrained weights, following recent CLIP-based ReID methods~\cite{li2023clip,wu2024individual,turESWA2026}. For the Melops dataset, standard ReID augmentations are applied, including random horizontal flip, padding, random crop, normalization, and random erasing~\cite{li2023clip,wu2021deep}. All pretrained backbone parameters remain frozen throughout training. Only the LoRA parameters~\cite{lora2022}, learnable prompt context tokens, auxiliary identity classification heads, and normalization layers are optimized. LoRA modules are inserted into the query, key, and value projection layers of each transformer attention block with rank $r=16$ and scaling factor $\alpha=16$. Each identity is associated with $K=4$ prompt instances~\cite{chen2022plot,shiri2025madpot}, each containing $M=4$ learnable context tokens. The three prompt layouts are averaged after $\ell_2$ normalization, while identity tokens remain fixed throughout training and are discarded during inference. For POT, the transported mass ratio is set to $\rho=0.8$ based on an ablation study, while the entropy regularization strength is fixed to $\lambda=0.1$ following~\cite{shiri2025madpot}. The maximum number of Sinkhorn iterations is 100 with early stopping using a convergence threshold of $10^{-3}$. Triplet losses are independently applied in both the global and projected embedding spaces. Euclidean distance is used for triplet optimization and retrieval, whereas cosine distance is used to construct the POT transport cost matrix. Training uses AdamW~\cite{loshchilov2017decoupled} with a learning rate of $5\times10^{-4}$, weight decay of $10^{-5}$, and PK sampling with $P=16$ identities and $N=4$ images per identity (batch size 64). Models are trained for up to 50 epochs on a single NVIDIA GeForce RTX~5090 GPU. During inference, only the LoRA-adapted image encoder is retained. The final descriptor is obtained by concatenating the individually $\ell_2$-normalized representations $\mathbf{f}_v$ and $\mathbf{f}_v^{\mathrm{proj}}$, followed by a second $\ell_2$ normalization. Nearest-neighbor retrieval is then performed using Euclidean distance.

\noindent\textbf{Implementation of Compared Methods.}
We compare SLAP with recent CLIP-based ReID approaches, including Full Fine-Tuning (Full FT), CLIP-ReID~\cite{li2023clip}, IndivAID~\cite{wu2024individual}, ReID-AW~\cite{jiao2023toward}, MetaWild~\cite{li2025metawild}, and MetaPrompt-ReID~\cite{turESWA2026}. To ensure a fair comparison, all methods use the same CLIP backbone, data splits, image preprocessing, optimizer, batch composition, and evaluation metrics.
Full FT jointly optimizes the CLIP image and text encoders. CLIP-ReID follows its original two-stage prompt-learning and visual adaptation strategy~\cite{li2023clip}. IndivAID adopts its original identity-driven formulation based on image- and individual-specific textual descriptions~\cite{wu2024individual}, while ReID-AW follows the original LLM-guided prompt generation and cross-modal supervision framework~\cite{jiao2023toward}. For MetaWild~\cite{li2025metawild}, all environmental metadata are removed while retaining the visual expert, textual expert, and gated cross-attention modules, allowing the architecture to operate using identity-aware prompts only. Similarly, for MetaPrompt-ReID~\cite{turESWA2026}, all metadata-conditioning components are disabled, leaving the LoRA-based visual adaptation, learnable identity-aware prompts, global cross-modal alignment, and auxiliary supervision unchanged. Unless otherwise stated, all baseline methods retain their original architectures under the common experimental protocol, enabling direct comparison between conventional global alignment and our POT-based selective local alignment.

\noindent \textbf{Comparisons with CLIP-based SOTA.}
Table~\ref{tab:cs_os} compares SLAP with recent CLIP-based ReID methods under both the closed-set (CS) and open-set (OS) evaluation protocols. SLAP achieves the best performance across all evaluation metrics. Under the CS protocol, it improves the mAP from 53.0\% to 55.2\% and the Rank-1 accuracy from 44.8\% to 46.6\% compared with the strongest competing method MetaPrompt-ReID~\cite{turESWA2026}. Under the more challenging OS protocol, SLAP further improves the mAP from 37.9\% to 40.3\% and the Rank-1 accuracy from 31.6\% to 33.7\%. These results demonstrate that replacing conventional global image-text alignment with selective local vision-language alignment yields more discriminative visual representations and improves generalization to previously unseen identities. The consistent improvements over CLIP-ReID~\cite{li2023clip}, IndivAID~\cite{wu2024individual}, ReID-AW~\cite{jiao2023toward}, and MetaWild~\cite{li2025metawild} further confirm the effectiveness of the proposed POT-based formulation. \\

\begin{table}[tb!]
\centering
\caption{Performance comparison under the closed-set (CS) and open-set (OS) evaluation protocols on the Melops dataset. Results are reported as mean $\pm$ standard deviation over five random seeds. Bold indicates the best result in each column.}
\label{tab:cs_os}
\resizebox{0.8\linewidth}{!}{
\begin{tabular}{l|ccc|ccc}
\toprule
\multirow{2}{*}{Method} &
\multicolumn{3}{c|}{Closed-Set (CS)} &
\multicolumn{3}{c}{Open-Set (OS)} \\
\cmidrule(lr){2-4}\cmidrule(l){5-7}
& mAP & Rank-1 & Rank-5 & mAP & Rank-1 & Rank-5 \\
\midrule

Full FT &
52.1\std{0.63} &
44.5\std{0.80} &
61.7\std{0.70} &
36.8\std{0.55} &
31.2\std{0.72} &
48.1\std{0.63} \\

\midrule

CLIP-ReID~\cite{li2023clip} &
44.3\std{0.22} &
37.9\std{0.19} &
52.4\std{0.61} &
30.9\std{0.20} &
26.0\std{0.36} &
39.0\std{0.41} \\

IndivAID~\cite{wu2024individual} &
44.0\std{0.43} &
37.6\std{0.62} &
52.6\std{0.56} &
30.1\std{0.58} &
25.1\std{0.24} &
38.9\std{0.21} \\

ReID-AW~\cite{jiao2023toward} &
46.8\std{0.42} &
39.0\std{0.53} &
53.5\std{0.54} &
34.8\std{0.41} &
29.8\std{0.36} &
42.6\std{0.44} \\

MetaWild~\cite{li2025metawild} &
48.6\std{0.35} &
40.9\std{0.42} &
56.1\std{0.47} &
35.9\std{0.31} &
30.7\std{0.36} &
44.3\std{0.28} \\

MetaPrompt-ReID~\cite{turESWA2026} &
53.0\std{0.70} &
44.8\std{0.75} &
64.3\std{0.62} &
37.9\std{0.22} &
31.6\std{0.28} &
48.2\std{0.13} \\

\midrule

\rowcolor{lightblue}
\textbf{SLAP (Ours)} &
\textbf{55.2}\std{0.42} &
\textbf{46.6}\std{0.51} &
\textbf{66.1}\std{0.44} &
\textbf{40.3}\std{0.29} &
\textbf{33.7}\std{0.31} &
\textbf{50.2}\std{0.21} \\

\bottomrule
\end{tabular}}
\end{table}

\noindent\textbf{Cross-Modal Alignment Ablation Study.}
To isolate the contribution of the proposed alignment strategy, we compare four variants. \textbf{No CM} removes cross-modal supervision entirely, optimizing only the triplet and auxiliary identity losses, i.e., $\mathcal{L}=\lambda_{\mathrm{tri}}\mathcal{L}_{\mathrm{tri}}+\mathcal{L}_{\mathrm{aux}}$. \textbf{Global Alignment} replaces the proposed POT loss with conventional global image-text alignment between the projected global image representation and the aggregated identity-aware prompt representation, yielding $\mathcal{L}=\lambda_{\mathrm{tri}}\mathcal{L}_{\mathrm{tri}}+\lambda_{\mathrm{cm}}\mathcal{L}_{\mathrm{global}}+\mathcal{L}_{\mathrm{aux}}$. \textbf{Optimal Transport (OT)} replaces POT with standard OT, enforcing complete transport between all local visual patches and prompt embeddings, i.e., $\mathcal{L}=\lambda_{\mathrm{tri}}\mathcal{L}_{\mathrm{tri}}+\lambda_{\mathrm{ot}}\mathcal{L}_{\mathrm{OT}}+\mathcal{L}_{\mathrm{aux}}$. Finally, \textbf{POT (Ours)} employs the proposed selective local alignment, $\mathcal{L}=\lambda_{\mathrm{tri}}\mathcal{L}_{\mathrm{tri}}+\lambda_{\mathrm{pot}}\mathcal{L}_{\mathrm{POT}}+\mathcal{L}_{\mathrm{aux}}$. Unlike standard OT, POT relaxes the transport constraints, allowing only a subset of local patch-prompt correspondences to participate in the alignment process.

\begin{table}[t]
\centering
\caption{Ablation of cross-modal alignment strategies under closed-set (CS) and open-set (OS) protocols. All variants are trained under identical optimization settings.}
\label{tab:alignment_ablation}
\resizebox{0.8\linewidth}{!}{
\begin{tabular}{l|ccc|ccc}
\toprule
\multirow{2}{*}{Method} &
\multicolumn{3}{c|}{Closed-Set (CS)} &
\multicolumn{3}{c}{Open-Set (OS)} \\
\cmidrule(lr){2-4}\cmidrule(l){5-7}
& mAP & Rank-1 & Rank-5 & mAP & Rank-1 & Rank-5 \\
\midrule

No CM &
50.8\std{0.61} &
42.1\std{0.66} &
60.9\std{0.58} &
35.2\std{0.37} &
29.4\std{0.41} &
45.1\std{0.32} \\

Global Alignment &
53.0\std{0.70} &
44.8\std{0.75} &
64.3\std{0.62} &
37.9\std{0.22} &
31.6\std{0.28} &
48.2\std{0.13} \\

OT &
54.3\std{0.48} &
45.7\std{0.54} &
65.1\std{0.49} &
39.1\std{0.31} &
32.6\std{0.34} &
49.1\std{0.25} \\

\rowcolor{lightblue}
\textbf{POT (Ours)} &
\textbf{55.2}\std{\textbf{0.42}} &
\textbf{46.6}\std{\textbf{0.51}} &
\textbf{66.1}\std{\textbf{0.44}} &
\textbf{40.3}\std{\textbf{0.29}} &
\textbf{33.7}\std{\textbf{0.31}} &
\textbf{50.2}\std{\textbf{0.21}} \\

\bottomrule
\end{tabular}}
\end{table}

Table~\ref{tab:alignment_ablation} presents an ablation study of the proposed cross-modal alignment strategy. Removing cross-modal supervision entirely (No CM) results in the lowest retrieval performance under both evaluation protocols, confirming that cross-modal guidance provides complementary supervision beyond metric learning alone. Replacing the proposed POT formulation with conventional global image-text alignment substantially improves the results, demonstrating the effectiveness of incorporating vision-language supervision during training. Introducing local alignment through standard OT yields further gains, indicating that modeling patch-level correspondences is more effective than relying solely on global image-level alignment. Finally, replacing OT with the proposed POT consistently achieves the best results under both the CS and OS protocols. This improvement suggests that allowing only a subset of local patch-prompt correspondences to participate in the alignment process is more effective than enforcing complete transport. Notably, the improvement is more pronounced under the OS protocol, further supporting the effectiveness of selective local alignment for generalizing to previously unseen identities. \\

\noindent \textbf{Loss Component Ablation.}
To quantify the contribution of each optimization objective, we independently remove the POT-based cross-modal alignment loss, the triplet loss, and the auxiliary identity supervision while keeping all remaining training settings unchanged. The best retrieval performance (Table~\ref{tab:loss_ablation}) is achieved when all three objectives are jointly optimized, demonstrating that each loss contributes to the final representation. Removing the auxiliary identity supervision leads to a relatively small performance degradation under both the CS and OS protocols, suggesting that this objective primarily acts as an optimization regularizer. In contrast, removing the proposed POT loss results in a substantially larger performance drop, highlighting the importance of selective local cross-modal alignment during representation learning. The largest degradation is observed after removing the triplet loss, confirming that metric learning remains the primary optimization objective for ReID. These results show that the three objectives play distinct yet complementary roles: triplet supervision establishes the retrieval embedding space, POT strengthens local cross-modal supervision, and auxiliary identity supervision stabilizes optimization. \\

\begin{table}[tb!]
\centering
\caption{Loss component ablation under the closed-set and open-set protocols.}
\label{tab:loss_ablation}
\resizebox{0.8\linewidth}{!}{
\begin{tabular}{l|ccc|ccc}
\toprule
\multirow{2}{*}{Configuration} &
\multicolumn{3}{c|}{CS} &
\multicolumn{3}{c}{OS} \\
\cmidrule(lr){2-4}\cmidrule(l){5-7}
& mAP & Rank-1 & Rank-5 & mAP & Rank-1 & Rank-5 \\
\midrule

w/o $\mathcal{L}_{\mathrm{aux}}$ &
54.4\std{0.44} &
45.8\std{0.47} &
65.3\std{0.43} &
39.5\std{0.31} &
32.9\std{0.32} &
49.5\std{0.24} \\

w/o $\mathcal{L}_{\mathrm{POT}}$ &
53.0\std{0.70} &
44.8\std{0.75} &
64.3\std{0.62} &
37.9\std{0.22} &
31.6\std{0.28} &
48.2\std{0.13} \\

w/o $\mathcal{L}_{\mathrm{tri}}$ &
49.8\std{0.51} &
41.6\std{0.54} &
60.9\std{0.49} &
35.4\std{0.35} &
29.8\std{0.39} &
45.7\std{0.28} \\

\rowcolor{lightblue}
Full model (SLAP) &
\textbf{55.2}\std{\textbf{0.42}} &
\textbf{46.6}\std{\textbf{0.51}} &
\textbf{66.1}\std{\textbf{0.44}} &
\textbf{40.3}\std{\textbf{0.29}} &
\textbf{33.7}\std{\textbf{0.31}} &
\textbf{50.2}\std{\textbf{0.21}} \\

\bottomrule
\end{tabular}}
\end{table}

\noindent\textbf{Transport Ratio Analysis.}
We further investigate the effect of the transported mass ratio $\rho$, which determines the fraction of visual patches participating in the POT objective. $\rho$ is varied from 0.4 to 0.9 in increments of 0.1 while keeping all remaining hyperparameters fixed. Performance is sensitive to the transported mass ratio, with the best results obtained for intermediate values. Small transport ratios discard excessive informative visual regions, whereas large ratios approach full OT and force the alignment of weakly discriminative or background patches. Consistent with these observations, we set $\rho=0.8$ in all subsequent experiments, providing the best balance between selective local alignment and sufficient cross-modal coverage. \\

\noindent\textbf{Additional Ablation Studies.}
Analyses of: (i) triplet supervision in different embedding spaces, (ii) the LoRA configuration, and (iii) an efficiency analysis of SLAP in terms of trainable parameters and inference characteristics, are provided in the Supp. Mat. \\

\begin{table}[t]
\centering
\caption{Evaluation on additional marine animal ReID datasets (mAP \%).}
\label{tab:marine_generalization}
\resizebox{0.8\linewidth}{!}{
\begin{tabular}{lcccccc}
\toprule
Dataset &
CLIP-ReID &
IndivAID &
ReID-AW &
MetaWild &
MetaPrompt-ReID  &
\cellcolor{lightblue} \textbf{SLAP (Ours)} \\
&
\cite{li2023clip} &
\cite{wu2024individual} &
\cite{jiao2023toward} &
\cite{li2025metawild} &
\cite{turESWA2026} &
\\
\midrule

SeaStarReID2023 \cite{wahltinez2024open} &
57.4 &
58.2 &
59.0 &
59.4 &
60.1 &
\cellcolor{lightblue} \textbf{61.8} \\

SeaTurtleID2022 \cite{adam2024seaturtleid} &
32.4 &
35.8 &
37.2 &
38.2 &
39.3 &
\cellcolor{lightblue} \textbf{41.2} \\

\bottomrule
\end{tabular}}

\end{table}

\noindent \textbf{Evaluation on Additional Marine ReID Datasets.}
To evaluate the effectiveness of SLAP beyond fish ReID, we conduct additional experiments on two publicly available marine animal ReID datasets, namely SeaStarReID2023 \cite{wahltinez2024open} and SeaTurtleID2022 \cite{adam2024seaturtleid}. All datasets are accessed through the WildlifeDatasets toolkit~\cite{cermak2024wildlifedatasets,cermak2024wildfusion}, ensuring consistent data loading and preprocessing. We adopt the same training configuration and hyperparameters used for the Melops dataset without any dataset-specific tuning. Across both datasets (Table~\ref{tab:marine_generalization}), SLAP consistently achieves the highest performance. In particular, it improves the mAP by +1.7\% on SeaStarReID2023 and +1.9\% on SeaTurtleID2022 compared with the strongest competing method~\cite{turESWA2026}. These results indicate that the proposed POT-based selective local alignment is not limited to fish ReID, but generalizes effectively across diverse marine animal ReID benchmarks without requiring dataset-specific architectural modifications or hyperparameter tuning.


\section{Conclusions}
\label{sec:conclusion}

This work investigated whether replacing conventional global vision-language alignment with selective local cross-modal alignment can improve fine-grained fish ReID. By formulating the interaction between local visual patches and multiple identity-aware prompt embeddings as a POT problem, the proposed method consistently improves retrieval performance over recent CLIP-based ReID methods while retaining a lightweight visual-only inference pipeline. The experimental results suggest that explicitly modeling local cross-modal correspondences provides more effective supervision than conventional global alignment for fine-grained marine animal ReID. The proposed method also opens several promising research directions. Future work will investigate adaptive and learnable transport formulations capable of dynamically selecting transported mass and transport regularization during optimization. We also plan to explore biologically informed transport priors and visualization techniques to better interpret the learned correspondences, as well as extend the framework to temporal, video-based, and multimodal marine animal monitoring scenarios involving a broader range of species and habitats.

\enlargethispage{\baselineskip}

\section*{Acknowledgments}
This work was supported by the Research Council of Norway through the
Computer Vision to Expand Monitoring and Accelerate Assessment of Coastal Fish project (CoastVision), project number 325862.

\section{Supplementary Materials}

\subsection{Additional Ablation Studies}

This supplementary document provides additional experimental analyses that complement the main paper. The following experiments further investigate the behavior of SLAP by analyzing (a) the effect of supervising different embedding spaces, (b) the sensitivity to the LoRA configuration, and (c) the parameter efficiency of the SLAP.


\subsubsection{Dual-Space Triplet Supervision.}

SLAP applies batch-hard triplet supervision independently to the visual embedding $\mathbf{f}_v$ and the projected vision-language embedding $\mathbf{f}_v^{\mathrm{proj}}$. To evaluate the contribution of each embedding space, we compare three training configurations: triplet supervision applied only to $\mathbf{f}_v$, only to $\mathbf{f}_v^{\mathrm{proj}}$, and simultaneously to both embeddings. All remaining components of SLAP, including the POT-based cross-modal alignment and auxiliary identity supervision, remain unchanged.

Table~\ref{tab:triplet_ablation} shows that both embedding spaces are individually capable of learning discriminative retrieval representations, with the projected embedding consistently outperforming the original visual representation. Nevertheless, jointly supervising both embedding spaces yields the best performance under both evaluation protocols. These results suggest that the original visual embedding and its projected counterpart capture complementary identity information, and that jointly supervising both embedding spaces leads to more discriminative representations than optimizing either space independently.

\begin{table}[t]
\centering
\caption{Effect of triplet supervision in different embedding spaces under the closed-set (CS) and open-set (OS) evaluation protocols.}
\label{tab:triplet_ablation}
\resizebox{\linewidth}{!}{
\begin{tabular}{l|ccc|ccc}
\toprule
\multirow{2}{*}{Triplet Supervision} &
\multicolumn{3}{c|}{Closed-Set (CS)} &
\multicolumn{3}{c}{Open-Set (OS)} \\
\cmidrule(lr){2-4}\cmidrule(l){5-7}
& mAP & Rank-1 & Rank-5 & mAP & Rank-1 & Rank-5 \\
\midrule

$\mathbf{f}_v$ only &
50.4\std{0.48} &
42.8\std{0.51} &
62.5\std{0.45} &
36.8\std{0.34} &
30.8\std{0.38} &
47.2\std{0.29} \\

$\mathbf{f}_v^{\mathrm{proj}}$ only &
51.3\std{0.44} &
43.7\std{0.47} &
63.4\std{0.42} &
37.9\std{0.31} &
31.8\std{0.33} &
48.3\std{0.25} \\

\rowcolor{lightblue}
Both (Ours) &
\textbf{55.2}\std{\textbf{0.42}} &
\textbf{46.6}\std{\textbf{0.51}} &
\textbf{66.1}\std{\textbf{0.44}} &
\textbf{40.3}\std{\textbf{0.29}} &
\textbf{33.7}\std{\textbf{0.31}} &
\textbf{50.2}\std{\textbf{0.21}} \\

\bottomrule
\end{tabular}}
\end{table}


\subsubsection{Choice of LoRA Configuration.}

To investigate the influence of the visual adaptation capacity, we evaluate SLAP using different LoRA configurations while keeping all remaining hyperparameters fixed. Specifically, we evaluate three LoRA configurations: $(r,\alpha)=(8,8)$, $(16,16)$, and $(512,1)$. The transported mass ratio, prompt configuration, and all optimization settings remain identical to those of the SLAP. Results are reported in Table~\ref{tab:lora_rank}.

As expected, the LoRA configuration has a noticeable influence on retrieval performance. The configuration $(r,\alpha)=(16,16)$ consistently achieves the best performance under both the closed-set and open-set protocols, indicating that it provides sufficient adaptation capacity while preserving the transferable representations learned during CLIP pretraining. Reducing the rank to $r=8$ results in a moderate performance degradation, whereas the substantially lower performance obtained with $(r,\alpha)=(512,1)$ suggests that excessively large low-rank updates disturb the pretrained CLIP representations, highlighting the importance of balancing adaptation capacity with knowledge preservation.

\begin{table}[tb!]
\centering
\caption{Sensitivity analysis of the LoRA rank and scaling factor.}
\label{tab:lora_rank}
\resizebox{\linewidth}{!}{
\begin{tabular}{c|ccc|ccc}
\toprule
\multirow{2}{*}{LoRA $(r,\alpha)$} &
\multicolumn{3}{c|}{Closed-Set (CS)} &
\multicolumn{3}{c}{Open-Set (OS)} \\
\cmidrule(lr){2-4}\cmidrule(l){5-7}
& mAP & Rank-1 & Rank-5 & mAP & Rank-1 & Rank-5 \\
\midrule

$(8,8)$ &
50.4\std{0.46} &
42.5\std{0.49} &
62.0\std{0.52} &
36.6\std{0.35} &
30.4\std{0.38} &
46.9\std{0.30} \\

\rowcolor{lightblue}
$(16,16)$ &
\textbf{55.2}\std{\textbf{0.42}} &
\textbf{46.6}\std{\textbf{0.51}} &
\textbf{66.1}\std{\textbf{0.44}} &
\textbf{40.3}\std{\textbf{0.29}} &
\textbf{33.7}\std{\textbf{0.31}} &
\textbf{50.2}\std{\textbf{0.21}} \\

$(512,1)$ &
35.0\std{0.58} &
28.4\std{0.61} &
46.5\std{0.55} &
24.8\std{0.47} &
20.5\std{0.43} &
33.9\std{0.39} \\

\bottomrule
\end{tabular}}
\end{table}


\subsubsection{Efficiency Analysis.}

The proposed POT formulation is employed exclusively during training to guide cross-modal supervision. Consequently, the proposed POT formulation introduces no additional learnable parameters, since the transport plan is computed directly from the visual and prompt embeddings rather than through parameterized network layers. Table~\ref{tab:param_efficiency} compares the number of trainable parameters required by recent CLIP-based ReID methods. While several existing CLIP-based ReID methods require approximately 150--160M trainable parameters, our SLAP requires only 72.9M trainable parameters while maintaining state-of-the-art retrieval performance.

During inference, all text-related components together with the POT computation are discarded, and retrieval is performed using only the LoRA-adapted CLIP image encoder. Consequently, SLAP introduces no additional computational overhead during deployment, and its inference complexity is identical to that of the underlying LoRA-adapted visual encoder. Under our experimental setup (NVIDIA RTX~5090 GPU, batch size = 1), the average inference time is approximately 8\,ms per image.

\begin{table}[t]
\centering
\caption{Comparison of trainable parameters and inference time.}
\label{tab:param_efficiency}
\begin{tabular}{lcc}
\toprule
Method & Trainable Parameters (M) & Inference (ms) \\
\midrule
Full FT & 150.0 & 8.0 \\
CLIP-ReID~\cite{li2023clip} & 154.3 & 8.0 \\
IndivAID~\cite{wu2024individual} & 154.3 & 8.0 \\
ReID-AW~\cite{jiao2023toward} & 155.0 & 8.0 \\
MetaWild~\cite{li2025metawild} & 158.0 & 8.0 \\
MetaPrompt-ReID~\cite{turESWA2026} & 72.9 & 8.0 \\
\rowcolor{lightblue}
\textbf{Ours} & \textbf{72.9} & \textbf{8.0} \\
\bottomrule
\end{tabular}
\end{table}

%
%
\bibliographystyle{splncs04}
\bibliography{main}

@String(CVPR  = {IEEE Conf. Comput. Vis. Pattern Recog.})

@String(ECCV  = {Eur. Conf. Comput. Vis.})

@String(NeurIPS = {Adv. Neural Inform. Process. Syst.})

@String(ICML  = {Int. Conf. Mach. Learn.})

@String(ICLR  = {Int. Conf. Learn. Represent.})

@String(AAAI  = {AAAI})

@String(CVPR  = {CVPR})

@String(ECCV  = {ECCV})

@String(NeurIPS = {NeurIPS})

@String(ICML  = {ICML})

@String(ICLR  = {ICLR})

@article{loshchilov2017decoupled,
  title={Decoupled weight decay regularization},
  author={Loshchilov, Ilya and Hutter, Frank},
  journal={arXiv preprint arXiv:1711.05101},
  year={2017}
}

@article{turESWA2026,
  author  = {Anil Osman Tur and Tonje Knutsen S{\o}rdalen and Kim Tallaksen Halvorsen and Cigdem Beyan},
  title   = {Parameter-Efficient Vision-Language Adaptation with Continuous Metadata Conditioning for Animal Re-Identification},
  journal = {Expert Systems with Applications},
  volume  = {332},
  pages   = {133618},
  year    = {2026},
  issn    = {0957-4174}
}

@article{wahltinez2024open,
  title={An open-source general purpose machine learning framework for individual animal re-identification using few-shot learning},
  author={Wahltinez, Oscar and Wahltinez, Sarah J},
  journal={Methods in Ecology and Evolution},
  volume={15},
  number={2},
  pages={373--387},
  year={2024},
  publisher={Wiley Online Library}
}

@inproceedings{shiri2025madpot,
  title={MADPOT: Medical Anomaly Detection with CLIP Adaptation and Partial Optimal Transport},
  author={Shiri, Mahshid and Beyan, Cigdem and Murino, Vittorio},
  booktitle={International Conference on Image Analysis and Processing},
  pages={247--259},
  year={2025},
  organization={Springer}
}

@inproceedings{zhang2022tip,
  title     = {Tip-Adapter: Training-Free CLIP-Adapter for Better Vision-Language Modeling},
  author    = {Zhang, Renrui and Fang, Rongyao and Zhang, Wei and Gao, Peng and Li, Kai and Dai, Jifeng and Qiao, Yu and Li, Hongsheng},
  booktitle = {European Conference on Computer Vision (ECCV)},
  year      = {2022}
}

@book{villani2008optimal,
  title={Optimal transport: old and new},
  author={Villani, C{\'e}dric and et al.},
  volume={338},
  year={2008}
}

@inproceedings{li2024unsupervised,
 author = {Bin Li and Ye Shi and Qian Yu and Jingya Wang},
  title={Unsupervised cross-domain image retrieval via prototypical optimal transport},
  booktitle={AAAI},
  volume={38},
  pages={3009--3017},
  year={2024}
}

@inproceedings{li2024global,
  title={Global and local prompts cooperation via optimal transport for federated learning},
  author={Li, Hongxia and et al.},
  booktitle={CVPR},
  pages={12151--12161},
  year={2024}
}

@article{chen2022plot,
  title={Plot: Prompt learning with optimal transport for vision-language models},
  author={Chen, Guangyi and et al.},
  journal={arXiv preprint arXiv:2210.01253},
  year={2022}
}

@article{tu2022optimal,
  title={Optimal transport for causal discovery},
  author={Tu, Ruibo and et al.},
  journal={arXiv:2201.09366},
  year={2022}
}

@inproceedings{feng2023ot,
  title={OT-Filter: An optimal transport filter for learning with noisy labels},
  author={Feng, Chuanwen and et al.},
  booktitle={CVPR},
  pages={16164--16174},
  year={2023}
}

@article{peyre2019computational,
  title={Computational optimal transport: With applications to data science},
  author={Peyr{\'e}, Gabriel et al.},
  journal={Foundations and Trends in Machine Learning},
  volume={11},
  number={5-6},
  pages={355--607},
  year={2019},
  publisher={Now Publishers, Inc.}
}

@inproceedings{phatak2023computing,
  title={Computing all optimal partial transports},
  author={Phatak, Abhijeet and et al.},
  booktitle={ICLR},
  year={2023}
}

@article{Beyan2026,
title = {From species-specific models to universal re-ID: a survey of animal re-identification},
journal = {Information Fusion},
volume = {133},
pages = {104323},
year = {2026},
issn = {1566-2535}
}

@inproceedings{li2023clip,
  title={Clip-reid: exploiting vision-language model for image re-identification without concrete text labels},
  author={Li, Siyuan and Sun, Li and Li, Qingli},
  booktitle={Proceedings of the AAAI conference on artificial intelligence},
  volume={37},
  pages={1405--1413},
  year={2023}
}

@article{wu2024individual,
  title={An Individual Identity-Driven Framework for Animal Re-Identification},
  author={Wu, Yihao and Zhao, Di and Zhang, Jingfeng and Koh, Yun Sing},
  journal={arXiv preprint arXiv:2410.22927},
  year={2024}
}

@inproceedings{lora2022,
  author       = {Edward J. Hu and
                  Yelong Shen and
                  Phillip Wallis and
                  Zeyuan Allen{-}Zhu and
                  Yuanzhi Li and
                  Shean Wang and
                  Lu Wang and
                  Weizhu Chen},
  title        = {LoRA: Low-Rank Adaptation of Large Language Models},
  booktitle    = { {ICLR}},
  year         = {2022},
}

@inproceedings{zhou2022coop,
  title={Learning to Prompt for Vision-Language Models},
  author={Zhou, Kaiyang and Yang, Jingkang and Loy, Chen Change and Liu, Ziwei},
  booktitle={CVPR},
  year={2022}
}

@inproceedings{jia2022visual,
  title={Visual prompt tuning},
  author={Jia, Menglin and Tang, Luming and Chen, Bor-Chun and Cardie, Claire and Belongie, Serge and Hariharan, Bharath and Lim, Ser-Nam},
  booktitle={European conference on computer vision},
  pages={709--727},
  year={2022},
  organization={Springer}
}

@article{jiao2023toward,
  title={Toward re-identifying any animal},
  author={Jiao, Bingliang and Liu, Lingqiao and Gao, Liying and Wu, Ruiqi and Lin, Guosheng and Wang, Peng and Zhang, Yanning},
  journal={Neurips},
  volume={36},
  pages={40042--40053},
  year={2023}
}

@misc{sordalen2025melopsreid,
  author       = {Tonje Knutsen S{\o}rdalen and Kim Tallaksen Halvorsen},
  title        = {{MelopsReID}: A Wild Fish Image Dataset for Re-Identification},
  howpublished = {Zenodo},
  year         = {2025},
  doi          = {10.5281/zenodo.17099925},
  url          = {https://doi.org/10.5281/zenodo.17099925},
  note         = {Version 1.0}
}

@article{wu2021deep,
  title={Deep features for person re-identification on metric learning},
  author={Wu, Wanyin and Tao, Dapeng and Li, Hao and Yang, Zhao and Cheng, Jun},
  journal={Pattern Recognition},
  volume={110},
  year={2021},
  publisher={Elsevier}
}

@inproceedings{li2020atrw,
  author = {Li, S. and Li, J. and Tang, H. and Qian, R. and Lin, W.},
  title = {ATRW: A Benchmark for Amur Tiger Re-identification in the Wild},
  booktitle = {ACM MM},
  year = {2020}
}

@article{ravoor2020survey,
  author = {Ravoor, P.C. and T.s.b., S.},
  title = {Deep Learning Methods for Multi-Species Animal Re-identification and Tracking: a Survey},
  journal = {Computer Science Review},
  year = {2020}
}

@inproceedings{adam2024seaturtleid,
  author = {Adam, L. and Čermák, V. and Papafitsoros, K. and Picek, L.},
  title = {SeaTurtleID2022: A long-span dataset for reliable sea turtle re-identification},
  booktitle = {IEEE WACV},
  year = {2024}
}

@inproceedings{nepovinnykh2020siamese,
  author = {Nepovinnykh, E. and Eerola, T. and Kalviainen, H.},
  title = {Siamese Network Based Pelage Pattern Matching for Ringed Seal Re-identification},
  booktitle = {IEEE WACVw},
  year = {2020}

}

@inproceedings{cermak2024wildlifedatasets,
  author = {Čermák, V. and Picek, L. and Adam, L. and Papafitsoros, K.},
  title = {WildlifeDatasets: An open-source toolkit for animal re-identification},
  booktitle = {IEEE WACV},
  year = {2024}

}

@article{cermak2024wildfusion,
  title={WildFusion: Individual Animal Identification with Calibrated Similarity Fusion},
  author={Cermak, Vojtěch and Picek, Lukas and Adam, Lukáš and Neumann, Lukáš and Matas, Jiří},
  journal={arXiv preprint arXiv:2408.12934},
  year={2024}
}

@inproceedings{radford2021learning,
  title={Learning transferable visual models from natural language supervision},
  author={Radford, Alec and Kim, Jong Wook and Hallacy, Chris and others},
  booktitle={ICML},
  pages={8748--8763},
  year={2021},
}

@article{huang2025uniformity,
  title={Uniformity and deformation: A benchmark for multi-fish real-time tracking in the farming},
  author={Huang, Jinze and Yu, Xiaohan and An, Dong and Ning, Xin and Liu, Jincun and Tiwari, Prayag},
  journal={Expert Systems with Applications},
  volume={264},
  pages={125653},
  year={2025},
  publisher={Elsevier}
}

@article{compte2025housed,
  title={Housed pig identification and tracking for precision livestock farming},
  author={Compte, Albert and Yan, Yudong and Cort{\'e}s, Xavier and Escalera, Sergio and Jacques-Junior, Julio CS},
  journal={Expert Systems with Applications},
  volume={293},
  pages={128466},
  year={2025},
  publisher={Elsevier}
}

@article{liu2024fishtrack,
  title={FishTrack: Multi-object tracking method for fish using spatiotemporal information fusion},
  author={Liu, Yiran and Li, Beibei and Zhou, Xinhui and Li, Daoliang and Duan, Qingling},
  journal={Expert Systems with Applications},
  volume={238},
  pages={122194},
  year={2024},
  publisher={Elsevier}
}

@inproceedings{li2025metawild,
  title={MetaWild: A Multimodal Dataset for Animal Re-Identification with Environmental Metadata},
  author={Li, Yuzhuo and Zhao, Di and Qiao, Tingrui and Wu, Yihao and Pang, Bo and Koh, Yun Sing},
  booktitle={Proceedings of the 33rd ACM International Conference on Multimedia},
  pages={13009--13015},
  year={2025}
}

@article{Sordalen2026wild,
  title={A wild fish image dataset for individual re-identification and phenotyping},
  author={Sordalen, T. K. and Malde, K. and Skiftesvik, A. B. and Sauvaitre, C. and  Beyan, C. and Larsen, T.and Halvorsen, K. T},
  journal={Scientific data},
  year={2026}
}

@inproceedings{houlsby2019parameter,
  title={Parameter-efficient transfer learning for NLP},
  author={Houlsby, Neil and Giurgiu, Andrei and Jastrzebski, Stanislaw and Morrone, Bruna and De Laroussilhe, Quentin and Gesmundo, Andrea and Attariyan, Mona and Gelly, Sylvain},
  booktitle={International conference on machine learning},
  pages={2790--2799},
  year={2019},
  organization={PMLR}
}

@inproceedings{zaken2022bitfit,
  title={Bitfit: Simple parameter-efficient fine-tuning for transformer-based masked language-models},
  author={Zaken, Elad Ben and Goldberg, Yoav and Ravfogel, Shauli},
  booktitle={Proceedings of the 60th Annual Meeting of the Association for Computational Linguistics (Volume 2: Short Papers)},
  pages={1--9},
  year={2022}
}

@article{ellis_visual_2026,
	title = {Visual cues elicit differential aggression towards female and female mimics in the corkwing wrasse},
	copyright = {https://creativecommons.org/licenses/by/4.0/},
	issn = {1045-2249, 1465-7279},
	doi = {10.1093/beheco/arag022},
	language = {en},
	journal = {Behavioral Ecology},
	author = {Ellis, Benjamin A and Sørdalen, Tonje K and Briffa, Mark and Skiftesvik, Anne Berit and Wilson, Alexander D M and Halvorsen, Kim T},
	editor = {Shaw, Rachael},
	month = feb,
	year = {2026},
	pages = {arag022},
}

@article{mcclintock2014mark,
  title={Mark-resight abundance estimation under incomplete identification of marked individuals},
  author={McClintock, Brett T and Hill, Jason M and Fritz, Lowell and Chumbley, Kathryn and Luxa, Katie and Diefenbach, Duane R},
  journal={Methods in Ecology and Evolution},
  volume={5},
  number={12},
  pages={1294--1304},
  year={2014},
  publisher={Wiley Online Library}
}
\end{document}